\documentclass[review]{elsarticle}

\usepackage{lineno,hyperref}
\usepackage{multirow}
\usepackage{amsmath}
\usepackage{amsfonts}
\usepackage{color}
\usepackage{array}
\newcommand{\PreserveBackslash}[1]{\let\temp=\\#1\let\\=\temp}
\newcolumntype{C}[1]{>{\PreserveBackslash\centering}p{#1}}
\newcolumntype{R}[1]{>{\PreserveBackslash\raggedleft}p{#1}}
\newcolumntype{L}[1]{>{\PreserveBackslash\raggedright}p{#1}}
\modulolinenumbers[5]

\journal{Pattern Recognition}

\begin{document}

\begin{frontmatter}

\title{Attention Driven Person Re-identification}

\author[pku]{Fan Yang\fnref{cor1}}
\author[ntu]{Ke Yan\fnref{cor1}}
\author[ntu]{Shijian Lu}
\author[pku]{Huizhu Jia\corref{cor2}}
\ead{hzjia@pku.edu.cn}
\author[pku]{Xiaodong Xie}
\author[pku]{Wen Gao}

\fntext[cor1]{The first two authors contributed equally to this work.}
\cortext[cor2]{Corresponding author.}

\address[pku]{National Engineering Laboratory for Video Technology, Peking University, Beijing, China}
\address[ntu]{Nanyang Technological University, 50 Nanyang Avenue, Singapore}

\begin{abstract}
Person re-identification (ReID) is a challenging task due to arbitrary human pose variations, background clutters, etc. It has been studied extensively in recent years, but the multifarious local and global features are still not fully exploited by either ignoring the interplay between whole-body images and body-part images or missing in-depth examination of specific body-part images. In this paper, we propose a novel attention-driven multi-branch network that learns robust and discriminative human representation from global whole-body images and local body-part images simultaneously. Within each branch, an intra-attention network is designed to search for informative and discriminative regions within the whole-body or body-part images, where attention is elegantly decomposed into spatial-wise attention and channel-wise attention for effective and efficient learning. In addition, a novel inter-attention module is designed which fuses the output of intra-attention networks adaptively for optimal person ReID. The proposed technique has been evaluated over three widely used datasets CUHK03, Market-1501 and DukeMTMC-ReID, and experiments demonstrate its superior robustness and effectiveness as compared with the state of the arts.
\end{abstract}

\end{frontmatter}

\section{Introduction}

Person re-identification (ReID) aims to identify the same individual across non-overlapping cameras. It has attracted increasing interests in recent years in the computer vision and pattern recognition research communities, largely due to its wide applications in surveillance analysis, etc. On the other hand, person ReID remains an open research challenge because of two major factors. First, the same person often has very large `intra-class' variation due to different imaging conditions in camera sensors, human poses, occlusion, background clutters and illuminations as illustrated in Fig. 1a. Second, as shown in Fig. 1b, the `inter-class' variation of different persons may be much smaller as compared with the `intra-class' variation of the same person. Most traditional methods address these challenges by either designing discriminative features \cite{Gray2008, Farenzena2010, Liu2012, Zhao2013, Zhao2014, Yang2014, Liao2015, Matsukawa2016} or learning powerful similarity metrics \cite{Prosser2010, ostinger2012, Li2013, Chen2016, Liao2015, Zhang2016, Yang2018TIP, WangPR2018, ZhouPR2017, RenPR2018}.

\begin{figure}
\centering
  \noindent\makebox[\linewidth][c] {
    \begin{minipage}[a]{0.08\linewidth}
        \centering
        \centerline{(a)}\medskip
    \end{minipage}

    \begin{minipage}[a]{0.75\linewidth}
        \centering
        \includegraphics[width=\linewidth]{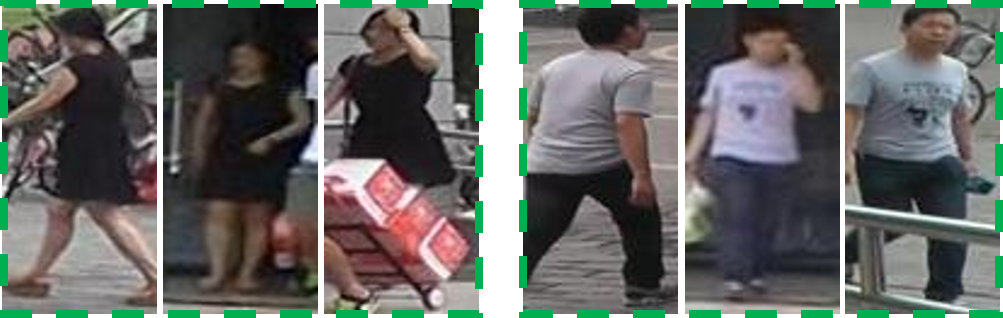}
    \end{minipage}
    }
  \noindent\makebox[\linewidth][c] {
    \begin{minipage}[b]{0.08\linewidth}
        \centering
        \centerline{(b)}\medskip
    \end{minipage}

    \begin{minipage}[a]{0.75\linewidth}
        \centering
        \includegraphics[width=\linewidth]{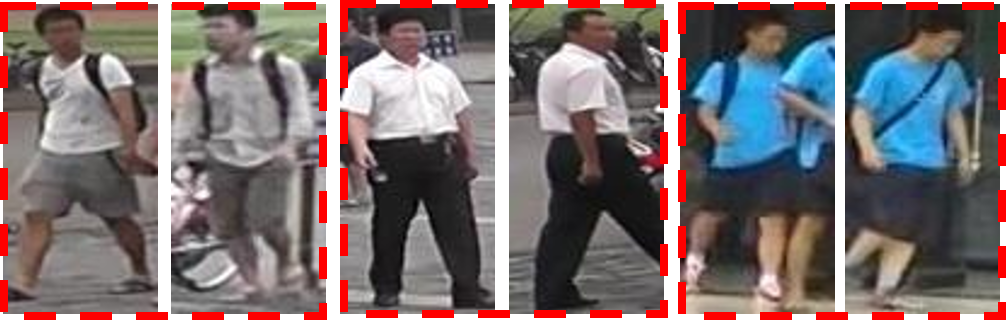}
    \end{minipage}
    }
  \noindent\makebox[\linewidth][c] {
    \begin{minipage}[b]{0.08\linewidth}
        \centering
        \centerline{(c)}\medskip
    \end{minipage}

    \begin{minipage}[a]{0.75\linewidth}
        \centering
        \includegraphics[width=\linewidth]{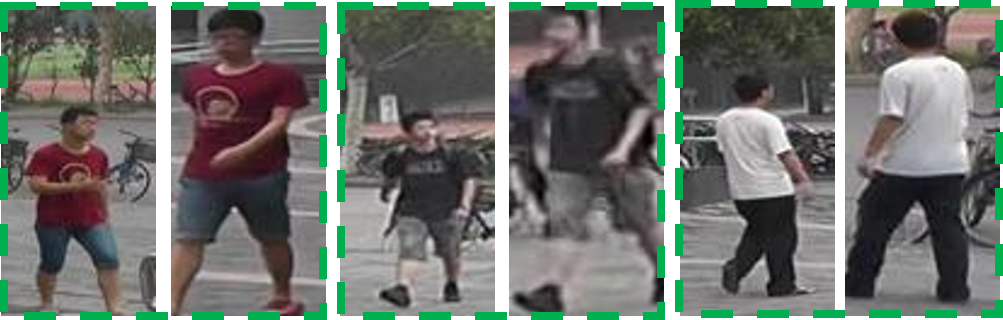}
    \end{minipage}
    }
\caption{Person ReID Challenges: The `intra-class' appearance variations of the same person in 1a (grouped by green-color boxes) may be larger than the `inter-class' appearance variations of different persons in 1b (grouped by red-color boxes) due to different human poses, occlusion, illuminations, etc. Additionally, large `intra-class' variations could be introduced by misalignment resulting from the  inaccurate human detection as illustrated in 1c.}
\label{fig:example}
\end{figure}

Deep neural networks have been widely used for the person ReID task in recent years. Leveraging large-scale person ReID datasets such as CUHK03 \cite{Li2014}, Market-1501 \cite{Zheng2015} and DukeMTMC-ReID \cite{Zheng_12017}, they have achieved very competitive performance and become prevalent in human visual feature learning. Most existing methods \cite{Xiao2016, Wang2016} learn a global representation from whole-body images but lose discriminative information lying around specific body parts. For example, the two distinct persons in the middle of Fig. 1b have very similar global appearance but fine differences around the head region. To capture the local discriminative information, several works \cite{Varior2016, Cheng2016} have been reported to learn part representations from some predefined horizontal partition strips. But human images collected by automatic detectors often suffer from misalignment and even part missing as illustrated in Fig. 1c. To address the misalignment issue, pose estimation \cite{Zhao2017, Wei2017} has been exploited to detect human parts to learn the local discriminative features. On the other hand, different regions within the same human part usually have different importance to the local discriminative feature learning, and different human parts also have different contributions to the final person ReID matching.

Visual attention can be exploited to detect informative pixels/regions within an image, which has good potential to train better deep network models by guiding the learning toward informative image regions \cite{Ba2015, Wang2017}. Given top-down target information, it helps to learn target relevant features and produces an attention map where regions of interest if present usually have much stronger response as compared with non-target regions. Attention has been used in person ReID, but most works \cite{Zhao_12017, Liu2017} learn global attention from whole-body images only where discriminative features of body-part images are often suppressed, e.g. the global attention has strong responses around certain specific regions only as illustrated in Fig. 2. Several attempts have been reported \cite{Zhao_12017, Li2018} to learn attention in multiple rounds under different parameters aiming to capture more local discriminative features, but the learnt attention is still global using whole-body images which often leads to redundant focus around similar regions.
\begin{figure}
\centering
\includegraphics[width=0.8\linewidth]{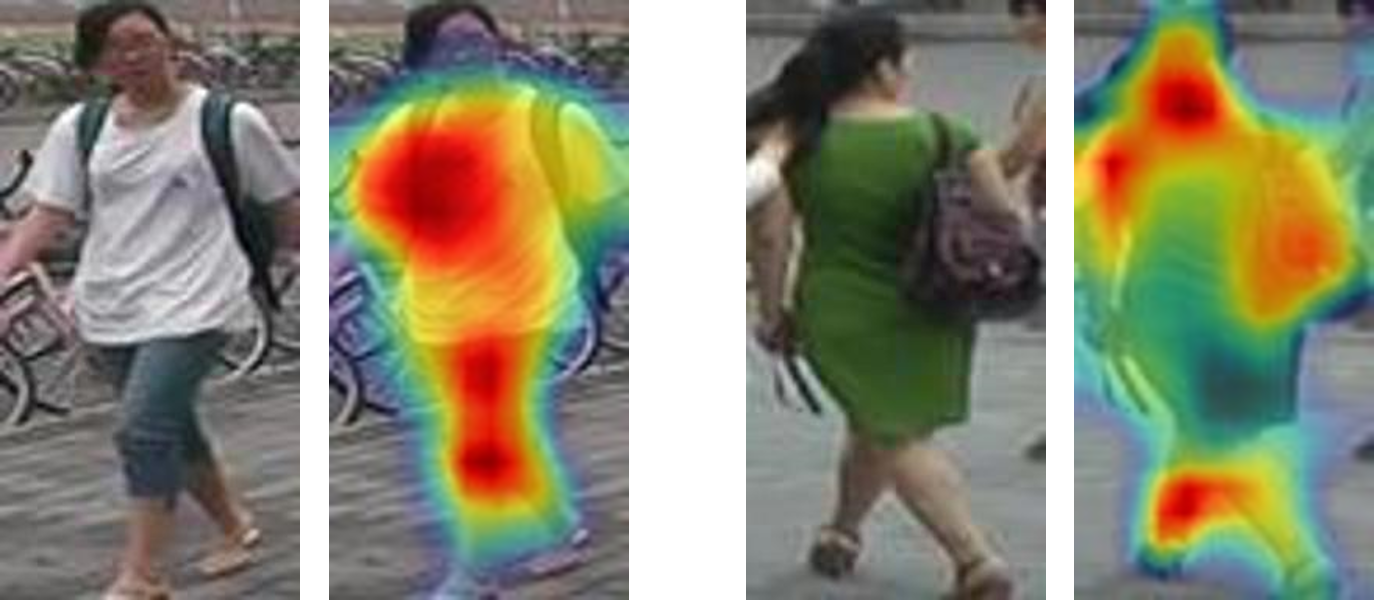}
\caption{Global attention detects globally discriminative regions which may suppress local informative regions and is insufficient in representation learning for person ReID by itself.}
\label{fig:heatmap}
\end{figure}

We design an attention-driven person ReID network that addresses the above constraints from two aspects: 1) it learns complementary discriminative representations from global whole-body images and local body-part images independently, and 2) it fuses the global and local features according to their learned contributions/importance to the feature matching. We formulate the two aspects by two specific terms, namely, intra-attention and inter-attention. The intra-attention aims to guide the learning to capture discriminative features of whole-body images and body-part images more precisely. For the whole body and each of the interested body parts, a dedicated intra-attention network is designed to learn the optimal feature representation and attention maps simultaneously. The inter-attention then learns optimal weights adaptively for optimal fusion of the output of intra-attention networks. To the best of our knowledge, this is the first attempt that models intra-attention and inter-attention under an end-to-end trainable network architecture. The proposed technique has four major contributions as listed:

\begin{itemize}
  \item It designs a novel multi-branch network architecture that learns precise and discriminative person ReID features under the guidance of intra-attention and inter-attention.
  \item It designs a novel intra-attention network that learns discriminative features from precisely aligned global whole-body images and body-part images concurrently and independently.
  \item It designs a novel inter-attention module that fuses discriminative features of the global whole-body images and local body-part images adaptively for optimal person ReID.
  \item It develops an end-to-end trainable deep network system that achieves superior person ReID performance across a number of widely used benchmarking datasets.
\end{itemize}

The rest of this paper is organized as follows. Section 2 briefly overviews related works. Section 3 presents our proposed method in detail. Implementation details and experimental results are presented in Section 4. Finally, several concluding remarks are drawn in Section 5.

\section{Related work}
Person ReID has been studied for years and a large number of person ReID techniques have been reported in the literature. This section will focus on prior works using deep networks because our proposed approach is deep network based and also deep network based techniques clearly outperform most prior works using `shallow' models. According to different learning strategies, existing deep network based methods can be broadly grouped into three categories including: 1) methods using global whole-body information only \cite{Ahmed2015, Li2014, Varior_12016, Zheng2016, Su2016, Xiao2016, Zheng2017, Su2017, Wang2016, Qian2017, Hermans2017, Chen2017, Sun2017, Lin2017, ZhouPR2018}, 2) methods using both global whole-body and local body-part information \cite{Yi2014, Varior2016, Shi2016, Cheng2016, Li2017, Zhao2017, Wei2017, Li_12017} and 3) methods using attention \cite{Wang2017, Ba2015, You2016, Liu2016, Zhao_12017, Liu2017, Li2018, Si2018, Chang2018}.

\subsection{Person ReID using whole-body information}
Earlier deep person ReID works learn global representation from whole-body images only. Different approaches have been reported to learn representation features and distance metrics by using different losses such as identity classification loss, pair-wise verification loss and triplet ranking loss \cite{Ahmed2015, Li2014, Varior_12016}. For example, Xiao et al. \cite{Xiao2016} train a classification model by treating images of a unique person as a specific category. In \cite{Zheng2017, Su2017}, pose-normalized images are included to train the classification model. Different networks have also been investigated for the person ReID problem. For example, Siamese networks which learn to estimate the similarity between a pair of images have been studied for person ReID by jointly considering classification and verification losses \cite{Zheng2016, Qian2017}. Triplet networks have also been studied for person ReID by learning relative similarity among three types of images including anchors, positive ones and negative ones. For example, Wang et al. \cite{Wang2016} combine triplet loss with a pairwise verification loss to unify a single-image representation and a cross-image representation and Hermans et al. \cite{Hermans2017} use a variant of the triplet loss to perform end-to-end deep metric learning. Quadruplet deep network which learn from four input images with a margin-based online hard negative mining strategy have also been investigated for person ReID problem recently \cite{Chen2017}. Further, different person attributes have been examined to improve the discrimination of the learned representation, e.g. Lin et al. \cite{Lin2017} explore complementary cues from attribute labels for better ReID performance and Su et al. \cite{Su2016} design a semi-supervised attribute learning framework to learn binary attribute features. Though these methods can learn global person representation effectively, they often produce sub-optimal ReID performance because they ignore the very informative details around body parts.

\subsection{Person ReID using whole-body and part information}
To address the constraints of using the whole-body information only, a number of new methods have recently been designed to capture richer and finer visual cues by jointly learning from both whole-body images and body-part images. These newly designed methods can be broadly classified into three categories depending on the part generation scheme. Methods in the first category use some predefined partition strategy such as fixed-height horizontal strips \cite{Varior2016, Cheng2016, Li2017}. This approach is simple to implement, but the predefined partitions are often poorly aligned with human body parts when human images are collected using imperfect automatic detectors. Methods in the second category use off-the-shelf pose estimation models to detect body parts. For example, Zhao et al. \cite{Zhao2017} first learn part representation and then fuse them with the global representation iteratively for person ReID. Wei et al. \cite{Wei2017} perform person ReID by concatenating part representation and global representation directly. Though learning representation from the estimated body parts alleviates the misalignment constraint, it can easily lead to failure when certain body parts are occluded or missing due to detection errors. Methods in the third category jointly learn part regions and features. For example, Li et al. \cite{Li_12017} propose to localize body parts using Spatial Transformer Networks (STN) \cite{J2015} but the learned body parts may belong to similar regions. Yao et al. \cite{Yao2017} propose to estimate body parts in a feature space and generate local features by ROI pooling, but the method is computationally intensive in both training and testing stages. More importantly, these methods focus more on developing robust part partition schemes instead of extracting discriminative features from body parts, and most of them ignore different importance while fusing local features of different body parts.

\subsection{Person ReID using attention}
In recent years, visual attention \cite{Li2016TIP} has been widely exploited to learn visual representations in various tasks in classification \cite{Wang2017, Zhang2018TIP}, object recognition \cite{Ba2015}, image captioning \cite{You2016} as well as person ReID \cite{Liu2012,Li2018,Liu2017,Si2018,Chang2018,ZhaoTPAMI2017}. Liu et al. \cite{Liu2012} propose an attention model that dynamically generates discriminative features from global whole-body images in a recurrent way. Li et al. \cite{Li2018} propose a Harmonious Attention (HA) model that locates body parts from whole-body images and learns multi-scale feature maps simultaneously. Liu et al. \cite{Liu2017} propose a multi-directional attention module that generates attentive features by masking different levels of features using an attention map. Si, et al. \cite{Si2018} extract feature vectors by pooling predefined sub-regions and then apply an intra-sequence attention mechanism to refine the extracted feature vectors. Chang et al. \cite{Chang2018} propose a Multi-Level Factorisation Net (MLFN) that learns visual factors at multiple semantic levels where an attentive factor selection module is designed to dynamically select which subset of factor modules are activated. Most existing works thus focus on attention learning using whole-body images, where dedicated attention learning from each body parts is largely neglected. On the other hand, global attention focuses more on global informative regions which often suppresses or ignores local informative regions around body parts and accordingly leads to suboptimal ReID performance when person images suffer from large pose variations, severe misalignments, local occlusion, etc.

The proposed attention-driven person ReID technique addresses the above constraints from two aspects. First, it learns complementary intra-attention from global whole-body images and local body-parts images independently. Second, it exploits inter-attention that fuses the global and local features according to their relevance to person ReID.

\section{Methodology}

Given $n$ training images $I = \{I_i\}_{i=0}^{n-1}$ of $q$ distinct person with the corresponding identity labels $L = \{L_i\}_{i=0}^{n-1}$ (where $L_i \in [0, ..., q-1]$), the target of person ReID is to learn a model that is capable of re-identifying images of the same person given some query image. We design a multi-branch attention-driven network that simultaneously learns and fuses discriminative and complementary features from both global whole-body images and local body-part images as illustrated in Fig. 3. The following subsections will describe the design and implementation of the attention-driven person ReID network in detail, specifically on the base network, the body part detection and alignment, the intra-attention network and the inter-attention module.

\begin{figure}
\centering
\includegraphics[width=0.9\linewidth]{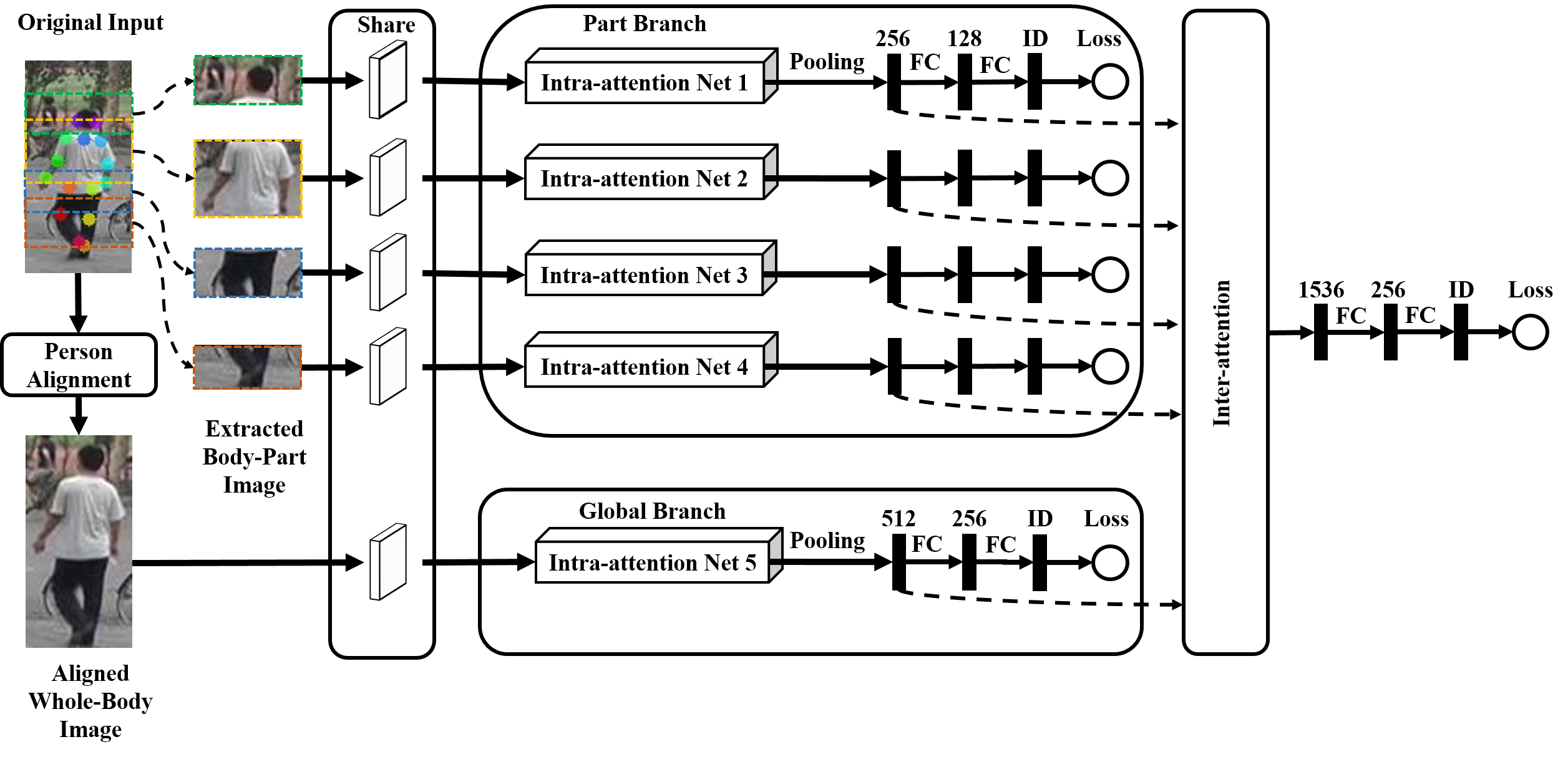}
\caption{The framework of the proposed attention-driven network: Given a person image $I_i$, an aligned whole-body image and four body-part images are first determined by pose estimation. Five intra-attention networks with shared lower convolutional layers then map the respective input image to discriminative features as supervised by independent softmax classification loss. An inter-attention module is further trained to fuse the outputs of five intra-attention networks according to their relevance to feature matching.}
\label{fig:framework}
\end{figure}

\subsection{Base network}
We adopt the Residual units \cite{He2016} as the basic building elements and design a multi-branch network for person ReID. As illustrated in Fig. 3, the designed network consists of five branches including: 1) one branch that aims to learn global features from whole-body images and 2) four independent branches that aim to learn local features from four body-part images. To minimize the model complexity, we simplify the ResNet50 in both network layers and channel numbers. In addition, we remove the last down-sampling operation in each branch (for higher granularity of the learnt features). All five branches learn independently, targeting optimal capture of complementary and discriminative identification features from whole-body images and four body-part images as well as minimization of overfitting risks. More details of the architecture of the base network are listed in Table 1.
\begin{table}[htb]
\caption{Detailed design and implementation of the base network: MP stands for max-pooling, AP stands for average-pooling and S stands for stride}
\centering
\scriptsize
\begin{tabular*}{0.8\linewidth}{@{\extracolsep{\fill}}|c|c|c|c|c|}  \hline
  Layer \# & Layer & Share & Global Branch & Part Branch\\  \hline  \hline
  1 & Conv1 & Yes & 3x3, 32, S-2x2 & 3x3, 32, S-2x2 \\  \hline
  \multirow{2}{*}{9} & \multirow{2}{*}{Conv2x} & \multirow{2}{*}{No} & 3x3 MP, S-2x2 & 3x3 MP, S-1x2 \\ \cline{4-5}
  & & &$\Bigg[\begin{tabular}{c}
              1x1, 32\\
              3x3, 32\\
              1x1, 64
            \end{tabular}\Biggl]$x3 &$\Bigg[\begin{tabular}{c}
              1x1, 16\\
              3x3, 16\\
              1x1, 32
            \end{tabular}\Biggl]$x3 \\  \hline
  9 & Conv3x & No &$\Bigg[\begin{tabular}{c}
              1x1, 64\\
              3x3, 64\\
              1x1, 128
            \end{tabular}\Biggl]$x3
            &$\Bigg[\begin{tabular}{c}
              1x1, 32\\
              3x3, 32\\
              1x1, 64
            \end{tabular}\Biggl]$x3 \\  \hline
  9 & Conv4x & No &$\Bigg[\begin{tabular}{c}
              1x1, 128\\
              3x3, 128\\
              1x1, 256
            \end{tabular}\Biggl]$x3
            &$\Bigg[\begin{tabular}{c}
              1x1, 64\\
              3x3, 64\\
              1x1, 128
            \end{tabular}\Biggl]$x3 \\  \hline
  9 & Conv5x & No &$\Bigg[\begin{tabular}{c}
              1x1, 256\\
              3x3, 256\\
              1x1, 512
            \end{tabular}\Biggl]$x3
            &$\Bigg[\begin{tabular}{c}
              1x1, 128\\
              3x3, 128\\
              1x1, 256
            \end{tabular}\Biggl]$x3 \\  \hline
  \multirow{2}{*}{1} & \multirow{2}{*}{FC\_reduce} & \multirow{2}{*}{No} & AP & AP \\ \cline{4-5}
  & & & 256  & 128  \\  \hline
  1 & FC & No & ID & ID \\  \hline
\end{tabular*}
\end{table}

\subsection{Body part detection and alignment}
Human images especially those collected using automatic detectors often suffer from various misalignment by either including too much background clutters or missing certain body parts. On the other hand, robust person ReID often requires good alignment of human body and sometimes even body parts. Based on the observation that human body and body parts can usually be localized by body joints, we employ pose estimation \cite{Cao2017, Yu2018TIE} to first localize body joints and then use the localized body joints for human alignment. In particular, we adopt an off-the-shelf pose estimator \cite{Cao2017} that directly produces 2D locations of 18 major body joints $K_j (j=1,...,18)$ as illustrated in Fig. 4a.

\begin{figure}
\centering
  \noindent\makebox[\linewidth][c] {
  \begin{minipage}[a]{0.15\linewidth}
  \centering
  \centerline{\includegraphics[width=\linewidth]{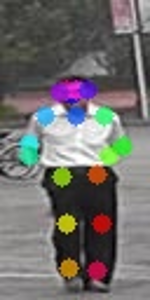}}
\end{minipage}
\begin{minipage}[a]{0.15\linewidth}
  \centering
  \centerline{\includegraphics[width=\linewidth]{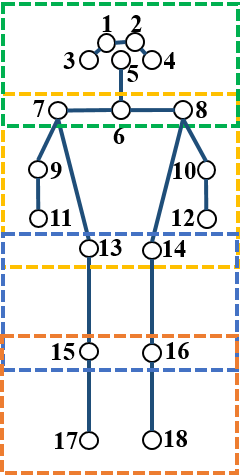}}
\end{minipage}
\begin{minipage}[a]{0.7\linewidth}
  \centering
  \centerline{\includegraphics[width=\linewidth]{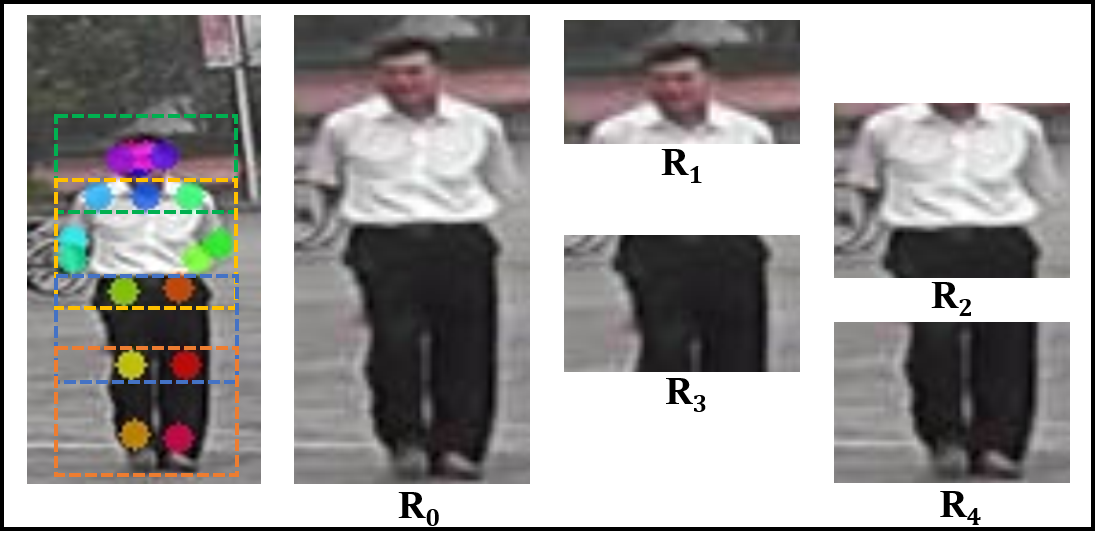}}
\end{minipage}
  }
  \noindent\makebox[\linewidth][c] {
  \begin{minipage}[a]{0.15\linewidth}
  \centering
  \centerline{(a)}\medskip
\end{minipage}
\begin{minipage}[a]{0.15\linewidth}
  \centering
  \centerline{(b)}\medskip
\end{minipage}
\begin{minipage}[a]{0.7\linewidth}
  \centering
  \centerline{(c)}\medskip
\end{minipage}
  }
\caption{Illustration of the human part detection: (a) 18 major human body joints by pose estimation, (b) definition of five body parts using the 18 major body joints, (c) detection of five body regions based on the definition in Fig. 4b.}
\label{fig:pose}
\end{figure}

Note that joint detection may suffer from detection errors while dealing with low-quality images due to occlusion, poor lighting, etc. We improve the joint detection by leveraging a set of canonical human poses that represent a list of typical human body configurations as exhibited in public surveillance cameras. Then for a new person image, only body joints with good detection confidence are kept and those missing or with ultra-low detection confidence are estimated by using the canonical poses. With 18 landmark body joints as illustrated in Fig. 4b, we first localize and divide the human images into five body regions as illustrated in Fig. 4c. In particular, the five regions are defined by the respective body joints which consist of the whole body region $P_0=\{K_1, ..., K_{18} \}$, the head region $P_1=\{K_1,...,K_8 \}$, the upper-body region $P_2=\{K_6,...,K_{14}\}$, the upper-leg region $P_3=\{K_{13},...,K_{16}\}$ and the lower-leg region $P_4=\{K_{15},...,K_{18}\}$. Let $(x_j, y_j)$ denote the coordinates of the 18 major body joints, several key parameters of the global human region can be computed as follows:
\begin{equation}
\begin{aligned}
& H = \frac{4}{3}\max \limits_{j \in P_4}(y_j) - \frac{1}{3}\min \limits_{j \in P_4}(y_j) - (2*\min \limits_{j \in P_1}(y_j)-\max \limits_{j \in P_1}(y_j)) \\
& (x_c, y_c) = (\overline{x_j}, \overline{y_j}) \quad j \in P_0 \\
& x_l = \min (x_c - H/4, \min \limits_{j \in P_0}(x_j)) \\
& x_r = \max (x_c + H/4, \max \limits_{j \in P_0}(x_j))
\end{aligned}
\end{equation}
where $(x_l, x_r)$, $(x_c, y_c)$ and $H$ denote the horizontal boundary, the center and the height of the whole body region, respectively. The bar symbol in $\overline{x_j}$ and $\overline{y_j}$ denotes a mean operator. For each body region $P_i, (i=0,...,4)$, the corresponding bounding box $R_i, (i=0,...,4)$ can thus be determined as follows:
\begin{equation}
R_i=
\begin{cases}
(x_l, 2*\min \limits_{j \in P_1}(y_j)-\max \limits_{j \in P_1}(y_j), x_r, \frac{4}{3}\max \limits_{j \in P_4}(y_j) - \frac{1}{3}\min \limits_{j \in P_4}(y_j)) \quad i=0, \\
(x_l, 2*\min \limits_{j \in P_1}(y_j)-\max \limits_{j \in P_1}(y_j), x_r, \max \limits_{j \in P_1}(y_j) + \beta) \quad i=1, \\
(x_l, \min \limits_{j \in P_i}(y_j)-\beta, x_r, \max \limits_{j \in P_i}(y_j)+\beta) \quad i =2,3 \\
(x_l, \min \limits_{j \in P_4}(y_j)-\beta, x_r, \frac{4}{3}\max \limits_{j \in P_4}(y_j) - \frac{1}{3}\min \limits_{j \in P_4}(y_j)) \quad i =4
\end{cases}
\end{equation}
where $\beta$ is the height of overlapping between two neighboring regions which is empirically set at $H/10$. For the sample image in Fig. 4a, Fig. 4c shows the determined five human regions.

\subsection{Intra-attention network}
The intra-attention network is constructed by stacking multiple representation learning blocks $B_i (i=1,...,4)$ as illustrated in Fig. 5. Multiple attention branches are adopted to generate attention maps at multiple resolutions, targeting to refine the learned representation progressively. In particular, features from the previous conv-layer are first fed into Block 1 to extract low level features. Blocks 2-4 each consists of two paths, one for feature extraction and the other for attention estimation. The feature extraction path acts as multiple detectors to extract semantic structures, where the initial features of the $i$-th block $B_i$ can be formulated as follows:
\begin{equation}
a_{i} = \mathcal{F}(v_i, \theta_i)
\end{equation}
where $v_i$ and $a_{i}$ are the input and output of feature extraction path of block $B_i$, and $\mathcal{F}$ is the stacked residual unit with parameters $\theta_i$. The output $a_{i}$ is a 3-D tensor $a_{i} \in \mathbb{R}^{h\times w \times c}$, where $h$, $w$, and $c$ denote the height, width, and channel number of the feature map $a_i$, respectively.

The attention path acts as a mask function to re-weight the features for automatic inference of regions of interest. It processes every input features of $B_i$ to obtain an attention score $m_i \in \mathbb{R}^{h\times w \times c}$ with the same size as $a_{i}$:
\begin{equation}
m_i = \mathcal{M}(v_i, \phi_i)
\end{equation}
where $\mathcal{M}$ is the attention scoring function with parameters $\phi_i$. With the attention mask, the basic form of the adjusted features become:
\begin{equation}
v_{i+1} = m_i \otimes a_{i}
\end{equation}
where $\otimes$ denotes element wise product, and $v_{i+1}$ is the output of $B_i$. Higher scores will be computed around the local regions that are more relevant to the the discriminative representation, largely driven by the loss function that aims to reduce the person ReID error to be described in Section 3.5.

More details of the proposed intra-attention network will be described in the following subsections, including encoder-decoder network, spatial-wise and channel-wise attention and optimization as illustrated in Fig. 5.
\begin{figure}
\centering
\includegraphics[width=0.9\linewidth]{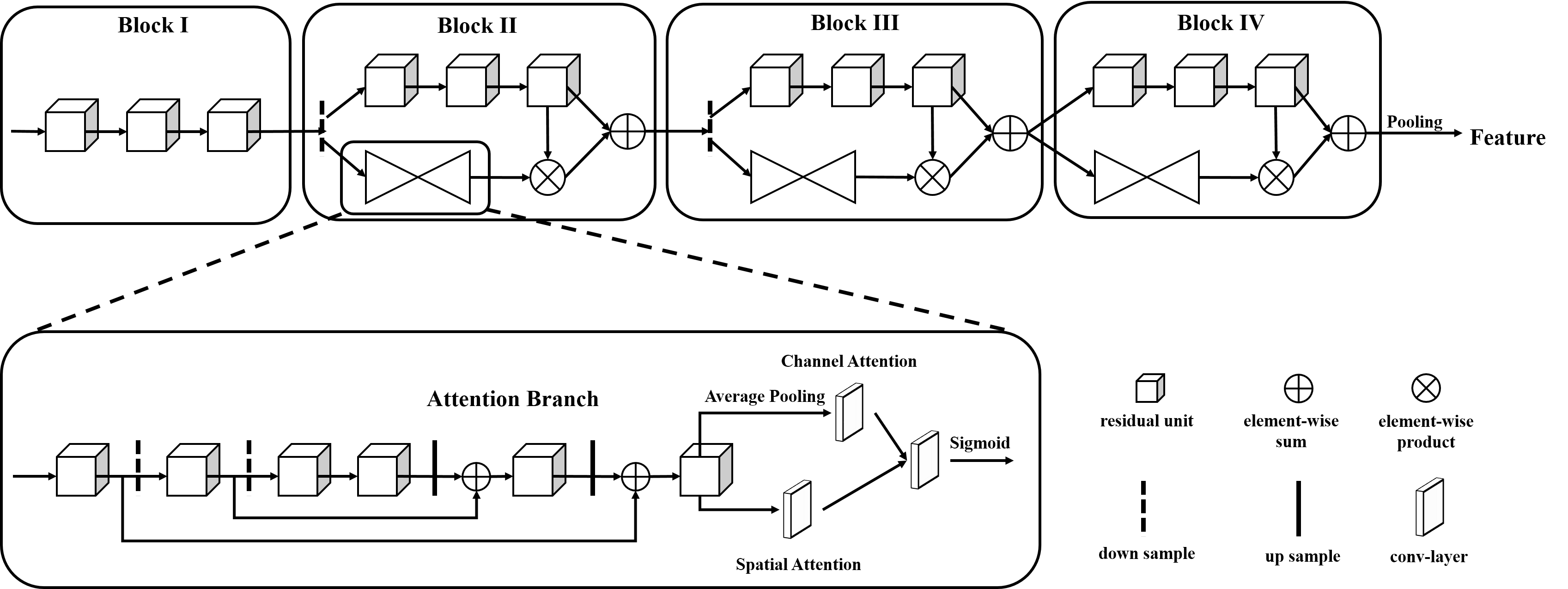}
\caption{Architecture of the proposed intra-attention network: multiple blocks are stacked to learn intra-attention at different scales. From the second block, an attention branch is included which learns spatial-wise attention and channel-wise attention simultaneously.}
\label{fig:intra}
\end{figure}

\noindent \textbf{Encoder-decoder network} Coherent understanding of the whole image and further focusing on discriminative local regions are essential for confidence estimation in various image recognition and classification tasks. For example, head, hat and glasses become the most discriminative visual cues around the head region while working on images of lower resolutions progressively. Aiming to capture discriminative features across multiple scales, we design an intra-attention network that employs the popular encoder-decoder structure as illustrated in Fig. 5. The encoder aims to learn multi-scale feature maps of the whole image region, where residual unit and max pooling are applied to process features down to lower resolutions. After reaching a predefined lowest resolution, the decoder employs symmetrical up-sampling iteratively to produce pixel-wise attention. To consolidate information across scales, skip layers are employed to combine features across the encoder and decoder at the same resolution, where the combination is implemented by an element wise addition of two sets of features. Note that we apply max pooling two times in Block 2, and one time in Blocks 3-4 as shown in Fig. 5.

\noindent \textbf{Spatial-wise and channel-wise attention} A convolutional layer employing $c$ channel filters scans an input image or feature map and outputs a $h\times w \times c$ feature map, where each filter detects one specific feature pattern across the spatial domain. Convolutional feature detection is therefore spatial-wise and channel-wise. Inspired by this observation, we design a novel attention learning strategy that decomposes attention into a spatial-wise attention component and a channel-wise attention component. Specifically, it decomposes attention of dimension $h\times w \times c$ (the same size as features as in conventional attention models) into a spatial-wise attention component of dimension $h\times w \times 1$ and a channel-wise attention component of dimension $1\times 1 \times c$. The decomposition thus reduces the number of parameters as well as the searching space significantly which helps to lower the model complexity and improve the person matching clearly, more details to be discussed in the experiments in Section 4.

In particular, the spatial-wise attention attempts to focus on semantic-related regions in the spatial domain, e.g. the head when we want to extract features from the head region as illustrated in Fig. 4c. In the proposed intra-attention networks, the spatial-wise attention $S_{i} \in \mathbb{R}^{h\times w \times 1}$ of block $B_i$ is computed by a convolutional layer which is formulated as follows:
\begin{equation}
S_{i} = f(A_i, W_i^s)
\end{equation}
where $f$ is the convolutional operation with parameter matrix $W_i^s$, $A_i$ is the initial attention confidence score as computed by the encoder-decoder network.

At the other end, each feature channel is actually an activation response of the corresponding convolutional filter and can be viewed as a semantic attribute. The learning of the channel-wise attention can therefore be interpreted as a process of selecting the most discriminative semantic attributes across multiple channels. For example, the channel-wise attention attempts to assign higher weights to the features of hat, glasses and hair while learning features around the head region. In implementation, an averaging pooling is first applied to each channel $A_i$ to obtain a channel feature $V_i \in \mathbb{R}^{1 \times 1 \times c}$. A convolutional layer is then employed to obtain the channel-wise attention map $C_i \in \mathbb{R}^{1\times 1 \times c}$. Finally, the spatial-wise attention $S_{i}$ and channel-wise attention $C_i $ are combined by first multiplying $S_i$ and $C_i$ and followed by a $1\times1$ convolution operation. The output $m_i$ is normalized to the range of $[0, 1]$ using a sigmoid function as illustrated in Fig. 5.

\noindent \textbf{Optimization} As studied in \cite{Wang2017}, element-wise production in Eq. 5 using a mask ranging from $0$ to $1$ may degrade the features of deep network layers. We address this issue by using a residual attention scheme which modifies the masking operation as follows:
\begin{equation}
v_{i+1} = (1 + m_i) \otimes a_{i}
\end{equation}
As defined in Eq. 7, the adjusted features will approximate the original ones when the attention score approximates 0. Otherwise, they are enhanced depending on the attention score. The new masking operation therefore attenuates the feature adjustment as compared with the one in Eq. 5.

\subsection{Inter-attention network}
The intra-attention networks learn discriminative features within the respective input images, where global features from whole-body images lay the groundwork and local features from body-part images capture complementary identification information. Local features of images of the four body parts usually capture different visual cues that have different contributions to the feature matching. Additionally, images of the four body parts often have different qualities due to variations in human poses, background clutters, etc. As a result, direct concatenation of the output of the four intra-attention body-parts networks with equal weights often leads to sub-optimal person ReID features.

We design an inter-attention module that adaptively fuses the output of the four intra-attention networks. Let $E_i(i=1,...,4)$ denote local features of the four intra-attention body-part networks, the corresponding weights $\mu_i$ can be learnt via four independent fully-connected layers as follows:
\begin{equation}
\mu_i = Sigmoid(w_iE_i + b_i)
\end{equation}
where $w_i$ and $b_i, i=1,...,4$ denote the weight vector and bias term of the four fully-connected layers. By applying the same aforementioned residual scheme in Eq. 7, the fused local feature $E_l$ can be derived by:
\begin{equation}
E_l = [(\mu_1+1)E_1, (\mu_2+1)E_2, (\mu_3+1)E_3, (\mu_4+1)E_4]
\end{equation}
where $\left[, \right]$ denotes concatenation.

The final person ReID feature is derived by assigning the same weight to the global feature $E_g$ (learnt from the whole-body images) and the fused part feature $E_l$. The principle here is that the global feature $E_g$ and the fused part feature $E_l$ have similar contributions to the person ReID though they usually capture complementary identification information, more details to be discussed in Section 4.4. Note we also investigated the scheme of learning the weight of $E_g$ in the similar way as in Eq. 8, but the obtained performance is slightly lower than the scheme described above.

\subsection{Loss functions}
We use the cross-entropy classification loss to train both intra-attention networks and inter-attention module. Given a training image $I_i$ with identity label $L_i$ and $X_i$ denoting the input of the prediction layer, the cross-entropy loss $l$ can be evaluated as follows
\begin{equation}
l = -\frac{1}{n} \sum_{i=0}^{n-1}log(\frac{exp(W_{L_i}X_i)}{\sum_{k=0}^{q-1}exp(W_{k}X_i)})
\end{equation}
where $n$ is the number of training images, $q$ is the number of identity and $W_k$ is parameter of the prediction function for the training identity $k$.

As described in the previous subsections, each intra-attention network is trained separately using an independent loss. Additionally, the inter-attention module also employs a loss to learn how to fuse features for optimal person ReID. The overall loss is thus defined as follows:
\begin{equation}
\mathcal{L} = \lambda\sum_{j=1}^{5}l_{intra}^j + l_{inter}
\end{equation}
where $\lambda$ controls the relative weights of the intra-attention loss and the inter-attention loss which is empirically set at 0.5 in our implemented system.

\section{Experiments}
\subsection{Datasets and settings}
\noindent \textbf{Datasets} Our proposed method is evaluated over three widely used datasets including CUHK03 \cite{Li2014}, Market-1501 \cite{Zheng2015} and DukeMTMC-ReID \cite{Zheng_12017} (Table 2). The CUHK03 consists of 14,097 images of 1,467 different identities, where 6 campus cameras were deployed for image collection and each identity is captured by 2 campus cameras. This dataset provides two types of annotations, one by manually labelled bounding boxes and the other by bounding boxes produced by an automatic detector \cite{Felzenszwalb2008}. The dataset also provides 20 random train/test splits used in \cite{Li2014} which selects 100 identities for testing and the rest for training. We select the first split and use 100 identities for testing and the rest 1,367 identities for training. The Market-1501 is collected using 6 cameras, which consists of 32,668 images of 1,501 identities as generated by an automatic detector. We follow the training and evaluation protocol in \cite{Zheng2015}, which splits images into a training set with 12,936 images and a testing set with 19,732 images. The DukeMTMC-ReID is a subset of DukeMTMC which was collected using 8 cameras for the study of cross camera tracking. It consists of images of 1,404 identities where half is used for training and the other half for testing. Specifically, there are 2,228 queries, 17,661 galleries, and 16,522 training images, respectively. We follow the protocol in \cite{Zheng_12017} for experiments on this dataset.

\begin{table}[h]
\caption{Settings of person ReID datasets that are used in the ensuing experiments.}
\centering
\begin{tabular}{|c|c|c|c|c|c|c}  \hline
  Dataset & Cams & IDs & Train IDs & Test IDs & Images\\  \hline
  CUHK03-Labeled  & 6 & 1,467 & 1,367 & 100 & 14,097   \\
  CUHK03-Detected  & 6 & 1,467 & 1,367 & 100 & 14,097   \\
  Market-1501  & 6 & 1,501 & 751 & 750 & 32,668   \\
  DukeMTMC-ReID  & 8 & 1,404 & 702 & 702 & 36,411   \\
  \hline
\end{tabular}
\end{table}

\noindent \textbf{Evaluation protocol}
The performance of person ReID is evaluated by using the widely used cumulative matching characteristic (CMC)\cite{KaranamTPAMI2018} across all three datasets. CMC is a widely used metric in person ReID evaluation. Take the single-gallery-shot setting (each gallery identity has only one instance) as an example. For each query, all gallery samples are ranked according to their distances to the query, and the CMC top-k accuracy is evaluated by:
\begin{equation}
Acc_k=
\begin{cases}
1 \quad \text{if top-k ranked gallery samples contain the query identity}, \\
0 \quad \text{otherwise}
\end{cases}
\end{equation}
It is actually a shifted step function, and the CMC curve can be derived by averaging the shifted step functions over all queries. Due to the space limit and also for direct comparison with the state of the arts, we only report the CMC accuracy at selected ranks instead of plotting actual CMC curves. The mean Average Precision (mAP) score \cite{Zheng2015} is also reported for the Market-1501 and DukeMTMC-ReID. But for the CUHK03, the mAP is not reported as in \cite{Li2014, Li_12017, Li2017, Su2017} because the gallery has only one image for each identity. All experiments adopt the single-query evaluation mode, and no re-ranking is performed for our method as well as compared methods.

\noindent \textbf{Implementation details}
Our ReID model is implemented and trained on the Keras \cite{Keras}, and Stochastic Gradient Descent (SGD) is used for optimization. The model is first pre-trained on the ImageNet (ILSVRC2012) \cite{Deng2009} for 9 epochs, where the learning rate is initially set as 0.01 and further divided by 10 after every 3 epochs. It is then fine-tuned on each of the three ReID datasets for 100 epochs, respectively, where the learning rate is initially set as 0.01 and further divided by 10 after every 40 epochs. The batch size is set at 32 for both pre-training and fine-tuning, and dropout is applied before every prediction layer with the dropout ratio empirically set at 0.5. Further, all training and testing image are rescaled to a fixed size of $384 \times 192$ and each of the four body parts has a fixed size of $96 \times 192$. Each training image is first normalized by subtracting its channel means and then fed to the network in a random order for training.

\subsection{Comparison with state of the arts}
The proposed method is evaluated and benchmarked with most state-of-the-art person Re-ID techniques over the three most widely used datasets including the Market-1501, the DukeMTMC-reID and the CUHK03.

For the Market-1501, the proposed method is compared with 17 state-of-the-art methods and Table 3 shows experimental results (the three methods above the horizontal line use traditional `shallow' model and the rest uses deep models). As Table 3 shows, our method achieves superior ReID accuracy and outperforms the state of the arts by $3.57\%$ in Rank-1 ($94.99\%$ versus $91.42\%$ by DuATM) and $9.85\%$ in mAP ($86.47\%$ versus $76.62\%$). Specifically, our method outperforms the pose-driven methods Spindle, PDC and GLAD (without using attention) by $18.09\%$, $10.95\%$ and $5.09\%$, respectively in Rank-1 and $21.8\%$, $23.06\%$ and $12.57\%$, respectively in mAP. The outstanding performance demonstrates the importance of using attention in feature learning. In addition, our method improves Rank-1 by $13.99\%$, $4.99\%$ and $3.79\%$ and mAP by $23.07\%$, $12.17\%$ and $10.77\%$, respectively, as compared to the Part-Aligned, MLFN and HA-CNN which use the global attention only. The clear performance gain is largely attributed to the intra-attention networks and inter-attention module that learn discriminative features from both global whole-body images and local body-part images simultaneously. Further, our strategy of progressive feature selection at multi-scale feature maps also helps to learn robust and discriminative features.
\begin{table}[h]
\caption{Comparison with the state of the arts on the Market-1501.}
\centering
\begin{tabular}{|c|ccccc|}
  \hline
  Methods & R1 & R5 & R10 & R20 & mAP \\
  \hline
  LOMO+XQDA \cite{Liao2015}   & 43.79 & - & - & - & 22.22 \\
  NFST \cite{Zhang2016}       & 55.43 & - & - & - & 29.87  \\
  BoW+Kissme \cite{Zheng2015} & 44.42 & 63.90 & 72.18 & 78.95 & 20.76  \\
  \hline
  HP-net \cite{Liu2017}          & 76.90 & 91.30 & 94.50 & 96.70 & -  \\
  Spindle \cite{Zhao2017}        & 76.90 & 91.50 & 94.60 & 96.70 & 64.67  \\
  MSCAN \cite{Li_12017}          & 80.31 & -     & -     & -     & 57.53  \\
  Part-Aligned \cite{Zhao_12017} & 81.00 & 92.00 & 94.70 & -     & 63.40  \\
  SVDNet \cite{Sun2017}          & 82.30 & -     & -     & -     & 62.10  \\
  PDC \cite{Su2017}              & 84.14 & 92.73 & 94.92 & 96.82 & 63.41  \\
  APR \cite{Lin2017}             & 84.29 & 93.20 & 95.19 & 97.00 & 64.67  \\
  TriNet \cite{Hermans2017}      & 84.90 & -     & -     & -     & 69.10  \\
  JLML \cite{Li2017}             & 85.10 & -     & -     & -     & 65.50  \\
  DPFL \cite{Chen_12017}         & 88.60 & -     & -     & -     & 72.60  \\
  GLAD \cite{Wei2017}            & 89.90 & -     & -     & -     & 73.90  \\
  MLFN \cite{Chang2018}          & 90.00 & -     & -     & -     & 74.30  \\
  HA-CNN \cite{Li2018}           & 91.20 & -     & -     & -     & 75.70  \\
  DuATM \cite{Si2018}            & 91.42 & 97.09 & -     & -     & 76.62  \\
  \hline
  Ours    & \textbf{94.99} & \textbf{98.25} & \textbf{99.12} & \textbf{99.38} & \textbf{86.47}\\
  \hline
\end{tabular}
\end{table}

For the larger and more recent dataset DukeMTMC-reID, the proposed method is compared with 9 state-of-the-art methods and Table 4 shows experimental results (the two methods above the horizontal line use traditional `shallow' model and the rest uses deep models). As Table 4 shows, our method obtains superior accuracy on a very different dataset, and it outperforms the state-of-the-art by $4.88\%$ in Rank-1 ($86.04\%$ versus $81.16\%$ by DuATM) and $6.84\%$ in mAP ($74.57\%$ versus $67.73\%$), respectively. This further verifies the advantages of our attention-driven network that employs intra-attention and inter-attention to guide feature learning and feature selection at multiple scales simultaneously. Note that lower accuracy is obtained over the DukeMTMC-reID as compared with the Market-1501, largely because images in the DukeMTMC-reID have more variations in image background and scene layout.
\begin{table}[!h]
\caption{Comparison with the state of the arts on the DukeMTMC-reID.}
\centering
\begin{tabular}{|c|cc|}  \hline
  Methods & Rank-1 & mAP \\  \hline
  LOMO+XQDA \cite{Liao2015}   & 30.75 & 17.04 \\
  BoW+Kissme \cite{Zheng2015} & 25.13 & 12.17  \\  \hline
  GAN(R) \cite{Zheng_12017}    & 67.68 & 47.13  \\
  APR \cite{Lin2017}           & 70.69 & 51.88  \\
  SVDNet \cite{Sun2017}       & 76.70 & 56.80  \\
  DPFL \cite{Chen_12017}      & 79.20 & 60.60  \\
  HA-CNN \cite{Li2018}        & 80.50 & 63.80  \\
  MLFN \cite{Chang2018}       & 81.00 & 62.80  \\
  DuATM \cite{Si2018}         & 81.16 & 67.73  \\
  \hline
  Ours                        & \textbf{86.04} & \textbf{74.57}\\
  \hline
\end{tabular}
\end{table}

For the CUHK03, two types of annotations are provided for each identity including manually labelled boxes and boxes produced by automatic detector. This dataset thus allows a direct model benchmarking in the presence of two types of most widely available annotations with distinct annotation quality. Table 5 show experimental results (the three methods above the horizontal line use traditional `shallow' model and the rest uses deep models). For the manually labelled boxes, our method outperform the state of the art by $4.63\%$ in Rank-1 ($96.43\%$ versus $91.80\%$ by HP-net). For the auto-detected boxes, our method wins out more, with a $10.78\%$ improvement in Rank-1 ($93.58\%$ versus $82.80\%$ by MLFN). This further shows the superior performance of our attention-driven network. By taking a second look, it can be observed that our model performs more consistently with respect to manually labelled and auto-detected boxes. It just obtains a $2.85\%$ improvement in Rank-1 for manually labelled boxes whereas most state-of-the-art methods have much larger performance drops while working with lower-quality boxes by automatic detector, e.g. 88.70\% versus 78.29\% by PDC, 86.70\% versus 82.00\% by DPFL, 85.40\% versus 81.60\% by Part-Aligned, etc.
\begin{table}[h]
\caption{Comparison with the state of the arts on the CUHK03 (CUHK03-L and CUHK03-D refer to the manually labelled boxes and auto-detected boxes).}
\centering
\begin{tabular}{|c|ccc|ccc|}  \hline
  \multirow{2}{*}{Methods} & \multicolumn{3}{c|}{CUHK03-L} & \multicolumn{3}{c|}{CUHK03-D}  \\ \cline{2-7}
  & R1 & R5 & R10 & R1 & R5 & R10 \\  \hline
  LOMO+XQDA \cite{Liao2015}        & 52.20 & 82.23 & 94.14 & 46.25 & 78.90 & 88.55  \\
  NFST \cite{Zhang2016}                & 58.90 & 85.60 & 92.45 & 53.70 & 83.05 & 93.00  \\
  GOG \cite{Matsukawa2016}            & 67.30 & 91.00 & 96.00 & 65.50 & 88.40 & 93.70     \\ \hline
  MSCAN \cite{Li_12017}                & 74.21 & 94.33 & 97.54 & 67.99 & 91.04 & 95.36  \\
  MuDeep \cite{Qian2017}              & 76.87 & 96.12 & 98.41 & 75.64 & 94.36 & 97.46     \\
  Part-Aligned \cite{Zhao_12017}     & 85.40 & 97.60 & 99.40 & 81.60 & 97.30 & 98.40  \\
  JLML \cite{Li2017}                      & 83.20 & 98.00 & 99.40 & 80.60 & 96.90 & 98.70  \\
  DPFL \cite{Chen_12017}              & 86.70 & -       & -       & 82.00 & -       & -          \\
  PDC \cite{Su2017}                      & 88.70 & 98.61 & 99.24 & 78.29 & 94.83 & 97.15  \\
  HP-net \cite{Liu2017}                  & 91.80 & 98.40 & 99.10 &  -     & -        & -        \\
  MLFN \cite{Chang2018}               & -       & -       & -       & 82.80 & -       & -           \\  \hline
  Ours                                         & \textbf{96.43} &\textbf{99.73} & \textbf{99.91} & \textbf{93.58} & \textbf{98.91} & \textbf{99.42} \\  \hline
\end{tabular}
\end{table}
\subsection{Ablation study}
Our proposed method learns discriminative person ReID features by using both global whole-body images and local body-part images. To tackle the misalignment and background clutters, pose estimation is employed to align the whole-body images and extract body-part images automatically. In addition, five branches of intra-attention networks are designed each of which learns attention of the whole body or one of four body parts, respectively. Further, an inter-attention module is designed which fuses the outputs of the five intra-attention networks according to their importance to person ReID. To find out how each of these innovative components help to achieve the outstanding person ReID performance in Tables 3-5, we develop five networks for ablation analysis including 1) a \textbf{baseline} model that implements the base multi-branch network without using attention (body parts are derived using predefined fixed horizontal strips \cite{Varior2016, Cheng2016}); 2) an \textbf{aligned} model that uses pose estimation to extract the four body parts beyond the \textbf{baseline}; 3) an \textbf{intra-attention} model that includes the intra-attention network in each branch beyond the \textbf{aligned} model; 4) an \textbf{inter-attention} model that includes the inter-attention module beyond the \textbf{aligned} model; and 5) an \textbf{intra+inter} model that include the inter-attention module beyond the \textbf{intra-attention} model.
\begin{table}[h]
\caption{Ablation study on the datasets Market-1501, DukeMTMC-reID and CUHK03 (CUHK03-L and CUHK03-D refer to the manually labelled and auto-detected boxes). *The R1 and mAP are evaluated by using optimal parameter $\beta$ and $\lambda$ to be discussed in the ensuing subsection `Parameter Setting'}
\centering
\begin{tabular}{|c|cc|cc|cc|cc|}  \hline
  \multirow{2}{*}{Models} & \multicolumn{2}{c|}{Market-1501} & \multicolumn{2}{c|}{DukeMTMC-reID} & \multicolumn{2}{c|}{CUHK03-L} & \multicolumn{2}{c|}{CUHK03-D}  \\
  \cline{2-9}  & R1 & mAP & R1 & mAP & R1 & mAP & R1 & mAP \\  \hline
  Baseline         & 86.63 & 66.25 & 75.98 & 57.23  & 85.46 & - & 80.02 & -\\
  Aligned             & 89.98 & 72.46 & 76.71 & 60.67  & 89.15 & - & 84.47 & -\\
  Intra   & 92.04 & 78.92 & 80.12 & 63.75  & 91.50 & - & 88.38 & -\\
  Inter   & 92.07 & 78.46 & 79.62 & 64.27  & 91.76 & - & 89.53 & -\\ \hline
  Intra+inter       & 94.65 & 85.22 & 84.78 & 71.92  & 94.96 & - & 93.45 & -\\
  *Intra+inter       & \textbf{94.99} & \textbf{86.47} & \textbf{86.04} & \textbf{74.57}  & \textbf{96.43} & - & \textbf{93.58} & -\\
  \hline
\end{tabular}
\end{table}

Table 6 show how the five networks perform over the three datasets where only Rank-1 and mAP results are shown. As Table 6 shows, the inclusion of pose estimation, intra-attention and inter-attention all helps to improve the person ReID performance clearly. The use of pose estimation consistently improves the person ReID performance, largely because it helps for more accurate person alignment and body part detection as compared with the use of some fixed predefined partitioning in the \textbf{baseline} model. In addition, either \textbf{intra} or \textbf{inter} model outperforms the \textbf{aligned} model consistently across the three datasets, demonstrating the effectiveness of using intra-attention and inter-attention in the person ReID problem. Furthermore, the concurrent inclusion of intra-attention and inter-attention in the \textbf{intra+inter} outperforms the use of either intra-attention or inter-attention alone, demonstrating the complementariness of the two proposed attention networks.
\begin{figure}
\centering
\includegraphics[width=0.9\linewidth]{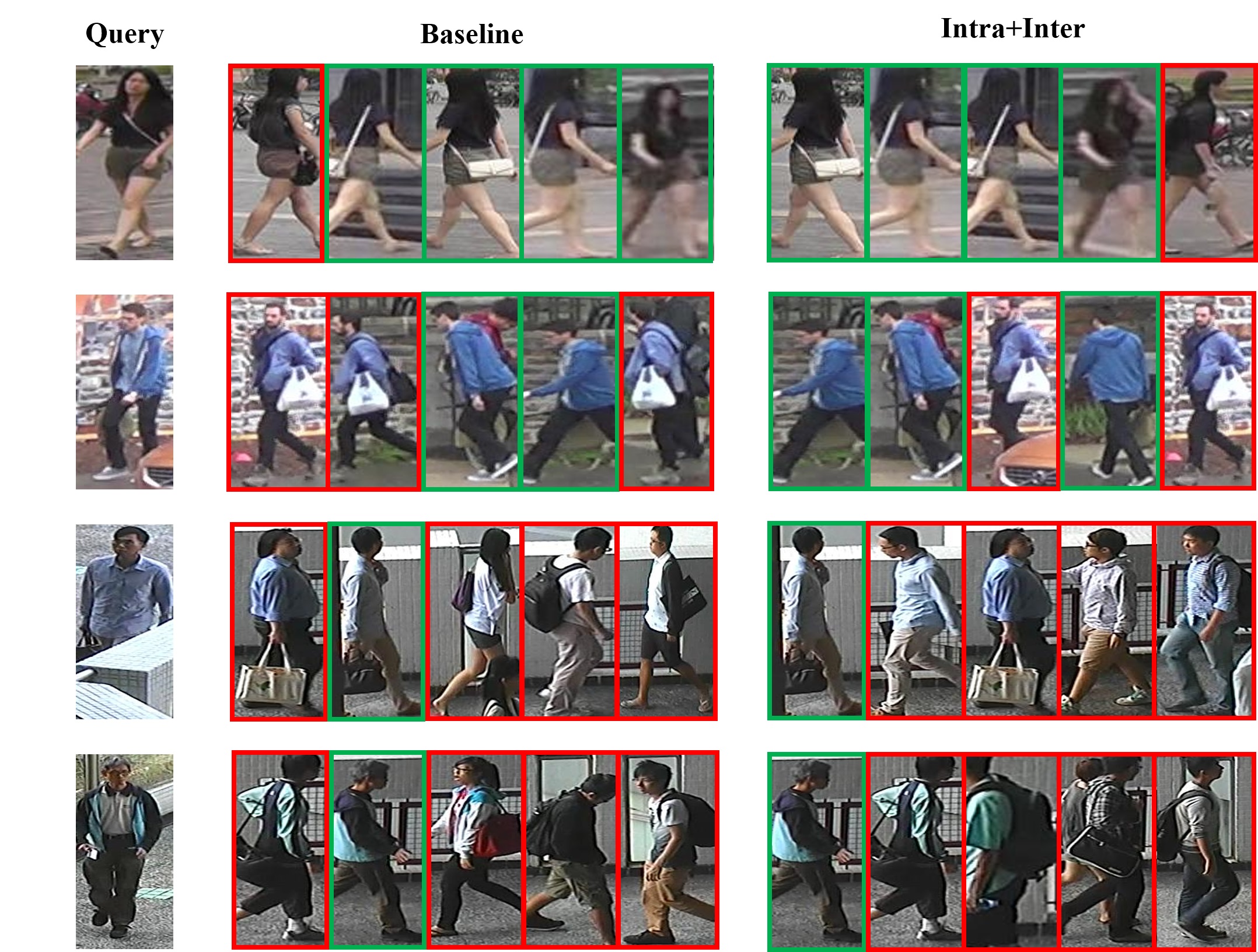}
\caption{Illustration of person ReID improvement using the proposed intra-attention and inter-attention: For the four sample images selected from the dataset Market-1501, DukeMTMC-reID, CUHK03-labeled and CUHK03-detected from top to bottom.For each query image in the first column, we compute its similarity to all gallery images and rank the gallery images according to their similarity to the query image. The second column and third column show the top five most similar gallery images as ranked by the `Baseline' and `Intra+Inter', where the green-color rectangles highlight person images which have the same ID with the query image and red-color rectangles highlight person images which have different ID from the query image.}
\label{fig:heatmap}
\end{figure}

Fig. 6 further illustrates how our proposed attention-driven network improves the \textbf{baseline} model that does not include pose-based alignment,  intra-attention and inter-attention. Four sample images are selected from the dataset Market-1501, DukeMTMC-reID, CUHK03 with manual human annotation and CUHK03 with automatic human detection. For each query image in the first column in Fig. 6, we compute its similarity to all gallery images and rank the gallery images according to their similarity to the query image. Fig. 6 shows the top five most similar gallery images as ranked by the `Intra+Inter' and the `Baseline', where the green-color rectangles highlight person images which have the same ID with the query image and red-color rectangles highlight person images which have different ID from the query image. In addition, the five images under both `Baseline' and `Intra+Inter' are five most similar gallery images that are arranged according to the similarity values from left to right. As Fig. 6 shows, the use of intra-attention and inter-attention helps to improve the person Re-ID performance significantly as compared with the \textbf{baseline} model.
\begin{table}[h]
\caption{Evaluations on how information from different human part contributes to the person ReID with and without using intra-attention (over the Market-1501)}
\centering
\begin{tabular}{|c|cccc|cccc|}  \hline
  \multirow{2}{*}{Market-1501} & \multicolumn{4}{c|}{With intra-attention} & \multicolumn{4}{c|}{Without intra-attention}  \\ \cline{2-9}
  & R1 & R5 & R10 & mAP & R1 & R5 & R10 & mAP \\  \hline
  Global            & 90.11 &96.05 &97.33 &75.23 &85.75 &94.54 &96.79 &68.24  \\  \hline
  Head              & 50.86 &73.31 &81.00 &26.76 &43.17 &67.37 &76.00 &21.83  \\
  Up-Body           & 48.96 &70.57 &78.06 &25.55 &43.37 &64.87 &72.74 &21.96  \\
  Up-Leg            & 48.96 &71.41 &79.54 &28.54 &42.66 &66.15 &75.38&23.54  \\
  Lower-Leg         & 36.22 &58.07 &67.51 &18.08 &30.58 &51.92 &61.01 &14.23 \\
  Part Fusion         & 91.62 &96.82 &98.01 &75.35 &87.10 &95.37 &97.62 &69.84\\ \hline
  Global+Part       & \textbf{94.65} &\textbf{98.21} &\textbf{98.96} &\textbf{85.22} &\textbf{92.07} &\textbf{97.35} &\textbf{98.51} &\textbf{78.46} \\
  \hline
\end{tabular}
\end{table}

\subsection{Discussion}
Beyond the ablation analysis, we also study three individual factors that could affect the person ReID performance including the contributions of individual human regions with and without using intra-attention, the decomposition of attention into spatial-wise attention and channel-wise attention, and different combinations of the local and global features.

\noindent \textbf{Intra-attention analysis} We evaluate how the global feature from whole-body images and local feature from body-part images contribute to the person ReID performance with and without intra-attention. Table 7 shows experimental results over the Market-1501 dataset. As Table 7 shows, global features from whole-body images have much higher contributions than local features from any individual body parts. In addition, the fusion of local features from the four body parts can achieve comparable performance with the global features. Furthermore, the combination of the global and local features further improves the performance with or without using the intra-attention. The different contributions of each body part also show the necessity of learning adaptive weights for feature fusion.
\begin{figure}
\centering
\includegraphics[width=0.9\linewidth]{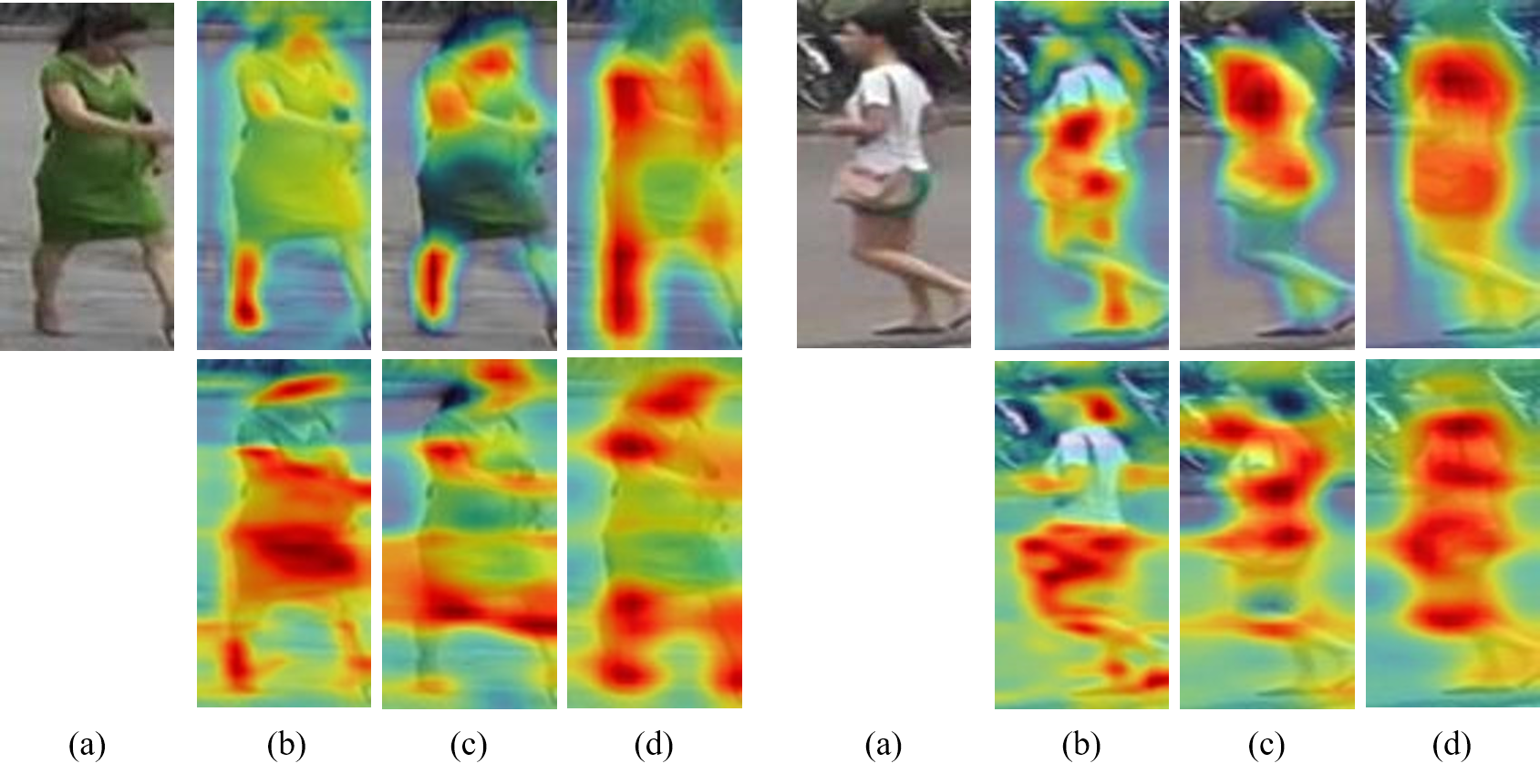}
\caption{Intra-attention of global whole-body images and local body-part images is complementary in feature selection: For each example image, the first row and the second row show three attention maps that are generated by global whole-body images and local body-part images, respectively. From left to right, (a) the original image, (b) the attention map from Block 2 in Fig. 5, (c) the attention map from Block 3, (d) the attention map from Block 4. Note the attention maps from images of four body parts are fused for better visualization.}
\label{fig:heatmap}
\end{figure}

The significance of using the intra-attention networks and inter-attention module to capture the complementariness of global features from whole-body images and local features from body-part images is further illustrated in Fig. 7. For each example image, six intra-attention maps are computed where the three in the first row are computed from global whole-body images and the three in the second row are computed from four body-part images. Additionally, each of the three intra-attention maps from left to right are output of Blocks 2, 3, and 4 in Fig. 5, respectively, which are computed progressively at different scales. As Fig. 7 shows, the intra-attention of whole-body images and body-part images are complementary which detects different regions for feature learning. In particular, the intra-attention of whole-body images detects more global structures whereas the intra-attention of body-part images detects more local details within respective body parts. This further shows the necessity and effectiveness of learning global and local level attention simultaneously.

\noindent \textbf{Spatial-wise and channel-wise attention} One key idea in the intra-attention networks is to first decompose the attention into spatial-wise attention and channel-wise attention and then derive the overall attention by multiplying the spatial-wise attention and channel-wise attention as described in Section 3.3. We study how this attention decomposition approach helps to improve the person ReID performance as compared with the traditional attention estimation without decomposition. Table 8 shows experimental results over the Market-1501 dataset. As Table 8 shows, the attention decomposition scheme improves the person ReID performance clearly, with a Rank-1 improvement by $1.39\%$ and a mAP improvement by $2.75\%$, respectively. This demonstrates the advantages and effectiveness of learning spatial-wise and channel-wise attention separately.
\begin{table}[h]
\caption{Comparison of attention estimation with and without decomposing into spatial-wise and channel-wise attention (over the Market-1501).}
\centering
\begin{tabular}{|c|ccccc|}  \hline
  Market-1501 & R1 & R5 & R10 & R20 & mAP \\  \hline
  Without attention decomposition   & 93.26 &97.92 &98.90 &99.17 &82.47   \\
  With attention decomposition     &\textbf{94.65} &\textbf{98.21} &\textbf{98.96} &\textbf{99.38} &\textbf{85.22}  \\
  \hline
\end{tabular}
\end{table}

\noindent \textbf{Inter-attention analysis}
We also study the effectiveness of the proposed inter-attention module by testing three feature fusion variants including: 1) direct concatenation of global and local features in testing stage; 2) conventional fully-connected fusion of global and local features in both training and testing stages; and 3) inter-attention fusion that learns adaptive feature weights. Table 9 shows experimental results over the Market-1501. As Table 9 shows, fusing features across global and local regions in training stage generally outperforms fusing features in testing stage only. In addition, fusing features using learned adaptive weights further improves the person ReID accuracy, which validates the rational of our inter-attention design that learning the relative importance of local features is beneficial to person ReID.
\begin{table}[h]
\caption{Comparison of different feature combination strategies over the Market-1501 (FC: fully-connected).}
\centering
\begin{tabular}{|c|ccccc|}  \hline
  Market-1501 & R1 & R5 & R10 & R20 & mAP \\  \hline
  Concatenation   & 89.98 &96.31 &97.68 &98.49 &72.46   \\
  FC fusion        & 90.94 &96.58 &97.89 &98.93 &77.11   \\
  Inter-attention  &\textbf{92.07} &\textbf{97.35} &\textbf{98.51} &\textbf{99.02} &\textbf{78.46}   \\
  \hline
\end{tabular}
\end{table}

\noindent \textbf{Parameter setting}
We first studied how $\lambda$ in loss function affects the person ReID performance over three datasets. The results are shown in Table 10, where two points can be observed: 1) a moderate $\lambda$ can bring extra supervision that helps to enhance the feature discriminability as learned by intra-attention networks; 2) the ReID performance is sensitive to $\lambda$ when person images have more occlusion and background clutters. The first point can be observed by the clearly higher person ReID accuracy when $\lambda$ becomes non-zero. The second point can be seen from the DukeMTMC-reID images that have wider camera views and more complex scene layout (as compared with Market-1501 and CUHK03 images) and so experience larger performance degradation when $\lambda$ becomes larger.
\begin{table}[h]
\caption{Evaluation on influence of parameter $\lambda$.}
\centering
\begin{tabular}{|c|cc|cc|cc|cc|}  \hline
  \multirow{2}{*}{$\lambda$} & \multicolumn{2}{c|}{Market-1501} & \multicolumn{2}{c|}{DukeMTMC-reID} & \multicolumn{2}{c|}{CUHK03-L} & \multicolumn{2}{c|}{CUHK03-D}  \\
  \cline{2-9}  & R1 & mAP & R1 & mAP & R1 & mAP & R1 & mAP \\  \hline
  0         & 83.64 & 63.70 & 72.17 & 54.35  & 80.22 & - & 74.83 & -\\
  0.1       & \textbf{94.98} & \textbf{86.45} & \textbf{85.90} & \textbf{74.10}  & 94.99 & - & 93.51 & -\\
  0.2       & 94.92 & 86.19 & 85.63 & 73.40  & \textbf{95.91} & - & \textbf{93.56} & -\\
  0.3       & 94.84 & 85.58 & 85.56 & 72.83  & 95.74 & - & 93.53 & -\\
  0.4       & 94.73 & 85.36 & 84.93 & 72.27  & 95.41 & - & 93.49 & -\\
  0.5       & 94.65 & 85.22 & 84.78 & 71.92  & 94.96 & - & 93.45 & -\\
  0.6       & 94.50 & 84.82 & 83.71 & 70.27  & 94.88 & - & 92.99 & -\\
  0.7       & 94.26 & 84.74 & 83.58 & 69.42  & 94.74 & - & 92.57 & -\\
  0.8       & 94.09 & 84.61 & 83.21 & 69.10  & 94.40 & - & 91.99 & -\\
  0.9       & 94.00 & 84.13 & 82.49 & 68.14  & 94.16 & - & 91.65 & -\\
  1.0       & 93.97 & 84.02 & 81.46 & 67.32  & 93.56 & - & 91.19 & -\\
  \hline
\end{tabular}
\end{table}

We also studied the impact of $\beta$, the height of the overlapping between two neighboring regions, over the Market-1501 and the DukeMTMC-reID. Experimental results are shown in Table 11, where we can see that the person ReID is not very sensitive to $\beta$.

In addition, the new studies also show that our proposed method achieves the best performance when $\beta$ and $\lambda$ are set at $(0.1, 3H/48)$, $(0.1, 3H/48)$ and $(0.2, 3H/48)$, for the Market-1501, DukeMTMC-reID and CUHK03 as shown in the last row of Table 6. The small variations of the optimal parameter settings across three very different datasets also demonstrate the robustness of our proposed technique.
\begin{table}[h]
\caption{Evaluation on influence of parameter $\beta$.}
\centering
\begin{tabular}{|c|cccc|cccc|}  \hline
  \multirow{2}{*}{$\beta$} & \multicolumn{4}{c|}{Market-1501} & \multicolumn{4}{c|}{DukeMTMC-reID}  \\ \cline{2-9}
  & R1 & R5 & R10 & mAP & R1 & R5 & R10 & mAP \\  \hline
  0            & 94.28 &98.14 &98.91 &85.19 &84.42 &92.23 &94.28 &71.26  \\
  $H/48$       & 94.35 &98.18 &98.81 &85.22 &84.87 &92.81 &94.66 &71.62  \\
  $2H/48$      & 94.36 &98.18 &98.78 &85.53 &85.41 &92.63 &\textbf{94.85} &71.87  \\
  $3H/48$      & \textbf{94.95} &\textbf{98.45} &\textbf{98.99} &\textbf{85.85} &\textbf{85.42} &\textbf{92.68} &94.48 &\textbf{72.32}  \\
  $4H/48$      & 94.77 &98.12 &98.87 &85.57 &85.14 &92.28 &94.32 &71.96  \\
  $5H/48$      & 94.65 &98.21 &98.96 &85.22 &84.78 &92.10 &94.30 &71.92  \\
  $6H/48$      & 94.53 &98.30 &98.90 &85.13 &84.41 &92.68 &94.79 &71.42  \\
  \hline
\end{tabular}
\end{table}

\noindent \textbf{Model complexity}
We compared the proposed model with four seminal classification CNN architectures (Alexnet \cite{Krizhevsky2012}, VGG \cite{Simonyan2015}, GoogLeNet \cite{Szegedy2015}, and ResNet50 \cite{He2016}) in model size and complexity. As the table 12 shows, our base network has the $2^{nd}$ smallest model size and the $2^{nd}$ smallest FLOPs, though it consists of five branches that share the first conv layer only. The fair model size and computational complexity is largely due to the simpler and smaller network model as presented in Table 1. Additionally, the intra-attention and inter-attention are both computational light and do not introduce much computational overhead.
\begin{table}[h]
\caption{Comparisons of model size and complexity. FLOPs: the number of FLoating-point OPerations; PN: Parameter Number.}
\centering
\begin{tabular}{|c|c|c|c|}  \hline
  Model & FLOPs & PN (million) & Stream \\  \hline
  AlexNet  & $1.07 \times 10^9$ &58.3 &1   \\
  VGG        & $2.28\times{10^{10}}$ &134.2 &1   \\
  ResNet50  & $5.58\times{10^9}$ & 23.5 & 1   \\
  GoogLeNet  & $2.31\times{10^9}$ & 6.0 & 1   \\
  \hline
  Basemodel  & $2.26\times{10^9}$ & 6.8 & 5   \\
  Full model  & $2.69\times{10^9}$ & 10.4 & 5   \\
  \hline
\end{tabular}
\end{table}

\section{Conclusion}
This paper proposes an end-to-end trainable network framework that learns a multi-branch attention-driven network model for accurate and robust person ReID. Different from most existing ReID methods that either ignore the matching misalignment problem or exploit global attention learning methods, the proposed intra-attention network is designed to detect informative regions within whole-body images and body-part images independently at multiple resolutions. This is achieved by the intra-attention module design in combination with a five-branch CNN architecture. In addition, a novel inter-attention module is designed which learns adaptive weights to fuse different intra-attention features for the optimal person ReID. Experiments over three widely used benchmarking datasets show that the proposed technique achieves superior person ReID performance as compared with the state of the art.

An ablation analysis is also performed to provide more insight of the designed network model. Leveraging our unique exploitation of local features of body-part images with inter and intra attention, we will continue to investigate more accurate and robust person ReID by incorporating trainable pose estimation and even semantic human part parsing in our future work.

\section{Acknowledgments}
This work is partially supported by National Key Research and Development Program of China under contract No. 2016YFB0402001, the Major National Scientific Instrument and Equipment Development Project of China under contract No. 2013YQ030967, National Science Foundation of China under contract No. 61602011 and NVIDIA NVAIL program. We would like to acknowledge supports from the Rapid-Rich Object Search (ROSE) Lab at Nanyang Technological University, Singapore.

\section*{References}

\bibliography{mybibfile}

\end{document}